\documentclass[runningheads,anonymous]{llncs}
\usepackage{comment}
\usepackage{graphicx}
\usepackage{amssymb} 
\usepackage{amsmath} 
\usepackage{multirow}
\usepackage{color}
\usepackage{pifont} 
\usepackage{hyperref}
\hypersetup{
    colorlinks=true,            
    linkcolor=black,             
    filecolor=magenta,          
    urlcolor=black,              
    pdftitle={Overleaf Example},
    pdfpagemode=FullScreen,
    }
\begin{document}
\title{Multimodal Distillation-Driven Ensemble Learning for Long-Tailed Histopathology Whole Slide Images Analysis}

\author{
Xitong Ling\inst{1}\textsuperscript{*} \and
Yifeng Ping\inst{2}\textsuperscript{*} \and
Jiawen Li\inst{1}\textsuperscript{*} \and
Jing Peng\inst{1} \and
Yuxuan Chen\inst{1} \and
Minxi Ouyang\inst{1} \and
Yizhi Wang\inst{1} \and
Yonghong He\inst{1} \and
Tian Guan\inst{1}\textsuperscript{\textdagger} \and
Xiaoping Liu\inst{3}\textsuperscript{\textdagger} \and
Lianghui Zhu\inst{1}\textsuperscript{\textdagger}
}
\begingroup
\renewcommand\thefootnote{\relax}
\footnotetext{* Contributed equally.}
\footnotetext{\textdagger\ Corresponding author.}
\endgroup
%

\authorrunning{Xitong Ling et al.}
\titlerunning{Multimodal Distillation-Driven Ensemble Learning for WSIs Analysis}
\institute{Shenzhen International Graduate School, Tsinghua University, China \\ 
\and
School of Interdisciplinary Studies, Lingnan university, China \\
\and
Zhongnan Hospital, Wuhan University, China \\
}
    
\maketitle              
\begin{abstract}
Multiple Instance Learning (MIL) plays a significant role in computational pathology, enabling weakly supervised analysis of Whole Slide Image (WSI) datasets. The field of WSI analysis is confronted with a severe long-tailed distribution problem, which significantly impacts the performance of classifiers. Long-tailed distributions lead to class imbalance, where some classes have sparse samples while others are abundant, making it difficult for classifiers to accurately identify minority class samples.
To address this issue, we propose an ensemble learning method based on MIL, which employs expert decoders with shared aggregators and consistency constraints to learn diverse distributions and reduce the impact of class imbalance on classifier performance. Moreover, we introduce a multimodal distillation framework that leverages text encoders pre-trained on pathology-text pairs to distill knowledge and guide the MIL aggregator in capturing stronger semantic features relevant to class information. To ensure flexibility, we use learnable prompts to guide the distillation process of the pre-trained text encoder, avoiding limitations imposed by specific prompts.
Our method, MDE-MIL, integrates multiple expert branches focusing on specific data distributions to address long-tailed issues. Consistency control ensures generalization across classes. Multimodal distillation enhances feature extraction. Experiments on Camelyon+-LT and PANDA-LT datasets show it outperforms state-of-the-art methods.

\keywords{Multiple Instance Learning \and  Long-Tailed Distribution \and Ensemble Learning.}

\end{abstract}
\section{Introduction}
Advances in digital imaging and computation have boosted computational pathology for diagnosis. However, annotating Whole Slide Images (WSIs) remains difficult due to their high resolution and specialized knowledge requirements.
Multiple Instance Learning (MIL) tackles these issues by using slide-level labels for tasks like cancer screening and classification. MIL treats WSIs as "bags" of patches, encodes patches into embeddings, and aggregates them for downstream analysis.

Previous MIL methods have primarily focused on two aspects. First, large-scale WSIs datasets are used for self-supervised pre-training of feature extractors. Examples include Ctranspath\cite{wang2022transformer}, UNI\cite{chen2024towards} and Gigapath\cite{xu2024whole}, which are pre-trained using pure image data. Additionally, multimodal feature encoders trained on image-text data using contrastive learning have been developed, such as PLIP\cite{huang2023visual}, CONCH\cite{lu2024visual}, and MUSK\cite{xiang2025vision}. Second, the design of aggregators for instance-level embeddings has evolved from prior-based non-parametric methods like Mean-Pooling and Max-Pooling to attention-based dynamic weight allocation mechanisms like ABMIL\cite{ilse2018attention}. The Transformer mechanism, which captures long-range dependencies and enables global modeling, has inspired many MIL methods based on self-attention, such as TransMIL\cite{shao2021transmil}, FR-MI\cite{10640165}, and AMD-MIL\cite{ling2024agent}. Other approaches include Graph Neural Networks (e.g., WiKG\cite{li2024dynamic}, HMIL\cite{chen2025dynamichypergraphrepresentationbone}), clustering methods (e.g., CLAM\cite{lu2021data}), pseudo-bag-based techniques (e.g., PMIL\cite{10663445}, SWS-MIL\cite{ouyang2024mergeup}) and data-augmentation (e.g., Norma\cite{10.1007/978-3-031-72983-6_24})

Enhanced feature extractors and aggregators have improved WSI analysis, but challenges persist, particularly due to class imbalance causing aggregator bias. Real-world diseases follow a long-tailed distribution, with rare diseases and common disease subtypes having low incidence rates. This distribution leads to classifiers overfitting to large-sample classes and underfitting to small-sample classes during deep neural network training, reducing overall performance and generalization on tail classes. Unlike lower-resolution datasets like natural images, CT, MRI, and OCT, WSIs have ultra-high resolution and unevenly distributed semantic information, making long-tailed analysis more difficult, especially under weakly supervised MIL labels. 

In natural image analysis, methods for addressing long-tailed distributions include class re-balancing, transfer learning, data augmentation, representation learning, and ensemble learning\cite{zhang2023deep}. Among these, ensemble learning shows potential in solving long-tailed problems by integrating expert models to learn various classes and make comprehensive decisions\cite{guo2021long,9156665}.
To address the long-tailed distribution problem in WSIs, we propose a multimodal distillation-driven ensemble learning (MDE-MIL) method. MDE-MIL uses a dual-branch structure to learn both the original long-tailed distribution and the balanced distribution after class re-balancing. WSIs are processed by a pre-trained feature extractor to generate patch embeddings, which are aggregated into slide-level embeddings by a shared-weight aggregator. These embeddings are fed into expert networks to output class probabilities optimized by cross-entropy loss. The shared-weight aggregator reduces parameters and maintains consistency, while Mean Squared Error (MSE) loss addresses potential inconsistencies between expert models.
To enhance the aggregator's ability to capture class-related information, we use a pathology multimodal pre-trained text encoder for knowledge distillation, aligning text and image embeddings and guiding the aggregator to perceive class semantics through label texts\cite{zhang2024vlm,li2024diagnostictextguidedrepresentationlearning}. Given the lack of text descriptions for WSIs and the challenges in generating captions, we introduce a learnable prompt initialized from a fixed-template prompt embedding. The prompt's average pooling result serves as the class text representation, aligned with slide representations for distillation, controlled by vector similarity and cross-entropy loss. To address the dimension inconsistency between MIL methods and text encoders, we use an MLP and a projection layer for alignment.
The contributions of this paper can be summarized as follows:

1. Our work alleviates the long-tailed distribution problem in MIL-based WSIs analysis, offering a new insight for research in this area.

2. We propose a novel multimodal distillation-driven ensemble learning method to tackle the long-tailed distribution problem in WSIs analysis.

3. We introduce a learnable prompt-based multimodal distillation mechanism to overcome limitations of transfer learning for WSI analysis.

4. Extensive experiments on the Camelyon$^{+}$-LT\cite{ling2024comprehensivebenchmarkpathologicallymph} and PANDA-LT\cite{bulten2022artificial} datasets demonstrate the superior performance of our method.

\section{Methodology}
\begin{figure}[t]
\includegraphics[width=\textwidth]{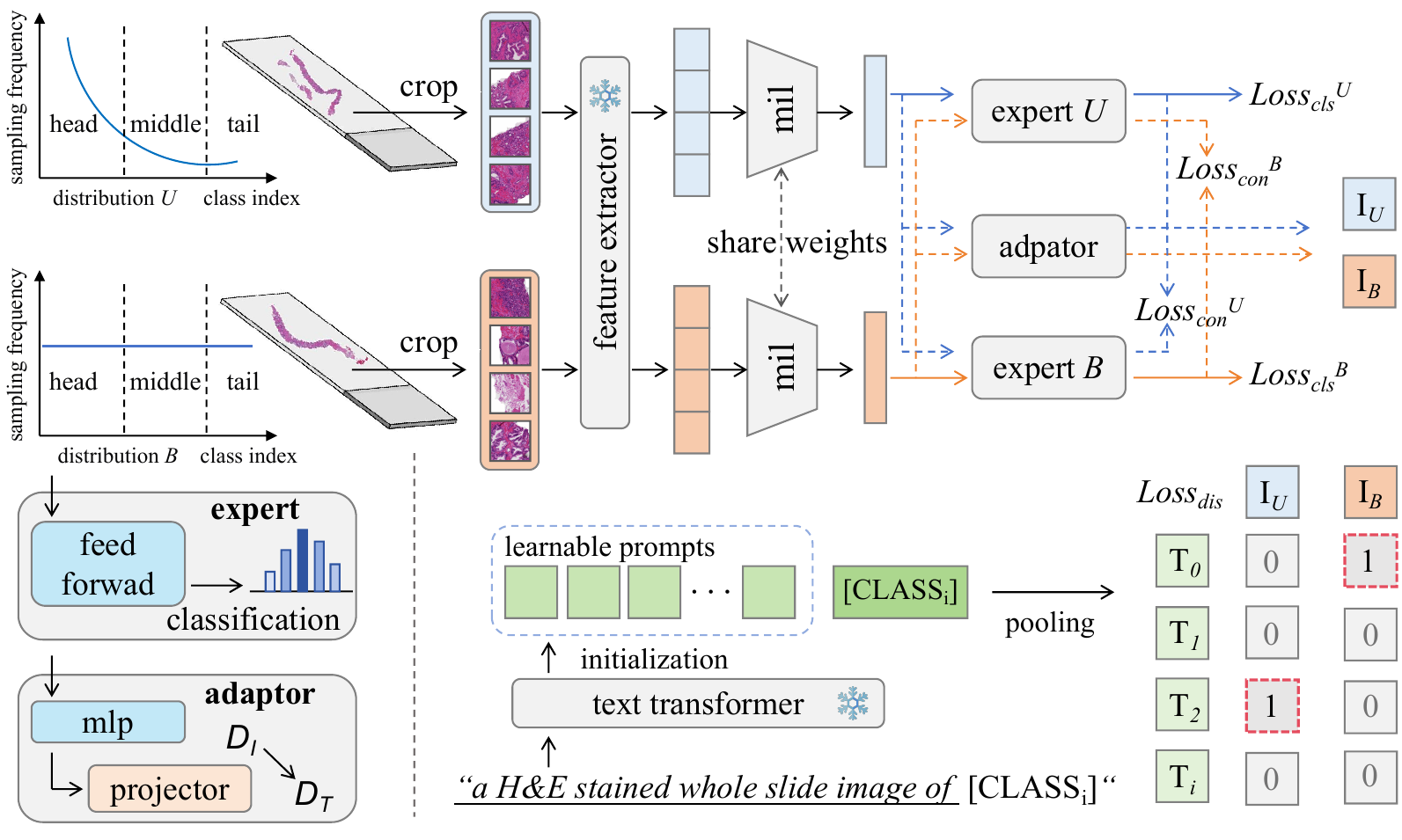}
\caption{The overall framework of MDE-MIL consists of an ensemble network and a multimodal distillation mechanism.} \label{fig1}
\end{figure}

In this section, we describe multiple instance learning in pathology and how MDE-MIL addresses the long-tailed effect in imbalanced data. The MDE-MIL architecture is shown in Figure \ref{fig1}.

\subsection{Problem Formulation}
Due to the exceptionally high resolution of WSIs, which typically encompass millions of local regions (patches), the weakly supervised MIL framework has emerged as one of the most prevalent paradigms for tasks such as lesion screening, tumor subtype classification, and gene mutation prediction.
In the MIL framework, each WSI is treated as a bag, and each patch is considered an instance.
Assume there are \( M \) WSIs, where the \( i \)-th WSI, \( X_i \), has label \( Y_i \) and contains \( N_i \) instances.
 Each instance \( x_{ij} \) (where \( j \in [1, N_i] \)) is first encoded by a pre-trained feature encoder \( f(\cdot) \) into a feature vector (embedding) of dimension \( d \):
\begin{equation}
x_{ij} \xrightarrow{f(\cdot)} e_{ij} \in \mathbb{R}^d
\end{equation}

Then, through an aggregator \( g(\cdot) \), the embeddings of all \( N_i \) instances in the \( i \)-th WSI are aggregated into a WSI-level embedding \( S_i \) of dimension \(D\):
\begin{equation}
S_i = g\left( \{ e_{i1}, e_{i2}, \dots, e_{iN_i} \} \right) \in \mathbb{R}^D
\end{equation}

Finally, through a classifier \( h(\cdot) \), the WSI-level embedding \( K_i \) is aligned with the downstream task for classification or prediction. Assume there are \( C \) categories, the classifier predicts the label \(\hat{y}_i\) (where \( \hat{y}_i \in \{ 0, 1, \dots, C-1 \} \)) of the \( i \)-th WSI:

\begin{equation}
\hat{y}_i = h(K_i)
\end{equation}

In the context of long-tail distribution, assume the \( k \)-th category contains \( M_k \) WSIs. The imbalanced-ratio \(r\) can be defined as:
\begin{equation}
r = \frac{\max(M_0, M_1, \dots, M_{C-1})}{\min(M_0, M_1, \dots, M_{C-1})}
\end{equation}

When \( r \) is large, the classifier \( h(\cdot) \) tends to favor the head classes, while the aggregator \( g(\cdot) \) also biases the aggregation of instances toward the prototype instances of the head classes. This will lead to a decrease in the performance of tail classes and affect the model's generalization ability.
\subsection{Ensemble Multiple Instance Learning}
The original unbalanced distribution of the dataset leads to class bias in both the aggregator and the classifier. Class rebalancing strategies can mitigate the long-tail effect. Suppose the original unbalanced distribution is $U$, where the probability of sampling each WSI in $U$ is $\frac{1}{M}$. The probability $P_U(X \in C_k)$ of sampling the WSI from the $k$-th class during training can be expressed as:

\begin{equation}
    P_U(X\in C_k) = \frac{M_k}{M}    
\end{equation}

Class rebalancing achieves the balanced distribution $B$ by weighting the sampling based on the inverse of the sampling probability for each class in the distribution $U$, The probability $P_B(X\in C_k)$ of sampling the WSI from the $k$-th class during training can be expressed as:
\begin{equation}
{P}_B(X\in C_k) = \frac{\frac{1}{M_k}}{\sum_{i=0}^{C-1} (\frac{1}{M_i} \times M_i)} \times M_k = \frac{1}{C}
\end{equation}

During training, in each iteration, samples $\{X_U, Y_U\}$ and $\{X_B, Y_B\}$ are drawn from distributions $U$ and $B$, respectively. These are passed through the pre-trained feature extractor $f(\cdot)$ to obtain the corresponding instance embeddings $\{ e_{U1}, e_{U2}, \dots, e_{UN_U} \}$ and $\{ e_{B1}, e_{B2}, \dots, e_{BN_B} \}$.
The distribution of instance counts representing class prototypes in some categories of WSI may be imbalanced, so using a dual-branch aggregator architecture may lead to poor intra-class consistency between aggregators. Additionally, aggregators based on transformer architectures and graph neural network architectures would increase the computational overhead. Therefore, we use a shared-weight aggregator to aggregate samples from different distributions.
Instance embeddings from different distributions are aggregated to WSI-level embeddings $S_U$ and $S_B$ through the weight-shared aggregator ${g_s}(\cdot)$:
\begin{equation}
\left\{
\begin{array}{l}
S_U = {g_s}\left( \{ e_{U1}, e_{U2}, \dots, e_{UN_U} \} \right), \\
S_B = {g_s}\left( \{ e_{B1}, e_{B2}, \dots, e_{BN_B} \} \right).
\end{array}
\right.
\end{equation}

Subsequently, $S_U$ and $S_B$ pass through two expert networks, $E_U(\cdot)$ and $E_B(\cdot)$, to obtain the class logits. Expert networks consist of FFN layers and classification heads. FFN provides a more expressive feature space through non-linear transformations, enhancing the network's ability to adapt to data from different distributions:
\begin{equation}
\left\{
\begin{array}{l}
z_U = E_U(S_U) \\
z_B = E_B(S_B)
\end{array}
\right.
\end{equation}

To ensure the intraclass consistency of different branch expert networks, \( S_U \) and \( S_B \) obtain their corresponding class logits \( \tilde{z_U} \) and \(\tilde{z_B} \) through \( E_B(\cdot) \) and \( E_U(\cdot) \), respectively. These logits are subsequently used for the consistency loss between different expert networks for the same input.

\subsection{Multimodal Knowledge Distillation}
The aggregator's attention to instances related to class semantics determines the quality of its representation of WSI information. However, due to the weak supervision provided by WSI labels, the aggregator's attention to key instances of minority classes is insufficient, thereby affecting its representation ability for these minority classes. Pathology multimodal pre-training based on contrastive learning aligns the text and image semantic spaces. Leveraging this property, textual labels of images serve as effective extractors of visual key instances and can thus guide the training of the aggregator. By distilling textual label knowledge into the aggregator, it can assign more weight to class-semantic instances, thereby achieving better aggregation results. We introduce an aggregator adaptor architecture to align the dimensions of slide embeddings and text embeddings. The adaptor consists of an MLP and a projector layer. The MLP enhances the representational capacity of slide embeddings, while the projector aligns the dimensions by mapping the slide embedding dimension \(D_I\) to the text embedding dimension \(D_T\).

In terms of text representation, we first construct a complete list of prompts using fixed template prompts and class-related text. Subsequently, we tokenize the complete list of prompts and use CONCH\cite{lu2024visual} to extract the text embedding \(S_T\). We then use the portion of the text embedding corresponding to the prompt template (\(S_{TP}\)) as initialization to create a trainable parameter \(S_{PL}\) of the same dimension, which is used to construct an automatically optimized prompt template. The portion of the text embedding representing class information (\(S_{TC}\)) is used as a fixed embedding and concatenated with \(S_{PL}\) to form the final prompt \(S_{TA}\). We compute the similarity between the average-pooled result of \(S_{TA}\) and the WSI embeddings from different distributions (\(S_U\) and \(S_B\)), and calculate the cross-entropy loss with their corresponding true labels to optimize the training.

\subsection{Optimization Strategy}
The optimization strategy of MDE-MIL mainly includes three loss functions: the classification loss \( \mathcal{L}_{\text{cls}} \) for controlling the classification of the two branches, the auxiliary consistency loss \( \mathcal{L}_{\text{con}} \) for maintaining consistency between the two branches, and the distillation loss \( \mathcal{L}_{\text{dis}} \) for text distillation. Among them, \( \mathcal{L}_{\text{cls}} \) and \( \mathcal{L}_{\text{dis}} \) use the cross-entropy loss function, while \( \mathcal{L}_{\text{con}} \) uses the MSE function. For the auxiliary loss functions, we introduce hyperparameters \( \alpha \) and \( \lambda \) to control their optimization weights. The total loss function \( \mathcal{L}_{\text{total}} \) can be expressed as:
\begin{equation}
    \mathcal{L}_{\text{total}} = \mathcal{L}_{\text{cls}} + \alpha \mathcal{L}_{\text{dis}} + \lambda \mathcal{L}_{\text{con}}
\end{equation}

\section{Experiment}
\subsection{Implementation Details}
We validate MDE-MIL's effectiveness on long-tailed problems using the Camelyon$^{+}$ and PANDA datasets. Camelyon$^{+}$ merges Camelyon16 and Camelyon17, containing 1350 slides (871 negative, 174 micro-metastasis, 251 macro-metastasis, 54 ITCs). PANDA-Karolinska has 5455 prostate biopsy slides divided into six classes (1924 G0, 1814 G1, 668 G2, 317 G3, 481 G4, 251 G5). We train on imbalanced datasets and test on balanced datasets. The Camelyon$^{+}$-LT split has an imbalance ratio of 35 , while PANDA-LT is 17 .
WSIs are divided into \( 256 \times 256 \) patches, and features are extracted using the CONCH\cite{lu2024visual}-pretrained extractor. Experiments are run on an RTX A6000 workstation with a base learning rate of 0.0002, weight decay of 0.00001, and Adam optimizer for 50 epochs, including a 2-epoch warmup. We use the same settings and data splits for fair comparisons. Experiments are repeated with five seeds, and we report the mean F1-score across head, medium, tail, and all classes.
\subsection{Main Results}
As shown in Table \ref{tab:mil_performance}, We compared our method with several state-of-the-art approaches, including Max-pooling, Mean-Pooling, ABMIL\cite{ilse2018attention}, Gate-ABMIL\cite{ilse2018attention}, CLAM\cite{lu2021data}, TransMIL\cite{shao2021transmil}, ADD-MIL\cite{NEURIPS2022_82764461}, FR-MIL\cite{10640165}, DS-MIL\cite{li2021dual}, DTFD\cite{zhang2022dtfd}, AMD-MIL\cite{ling2024agent}, and ILRA-MIL\cite{Xiang2023ExploringLP}. MDE-MIL achieved significant improvements in metrics for both tail classes and overall classes compared to the baseline models, thereby demonstrating the effectiveness of the MDE architecture.

\begin{table*}[h]
    \centering
    \setlength\tabcolsep{4pt} 
    \caption{Performance comparison of different MIL methods on various datasets.}
    \begin{tabular}{c|cccc|cccc}
        \hline
        \multirow{2}{*}{\textbf{Method}} & \multicolumn{4}{c|}{\textbf{Camelyon$^{+}$-LT (\(r\) = 35)}} & \multicolumn{4}{c}{\textbf{PANDA-LT (\(r\) = 17)}} \\
        & Head & Medium & Tail & All & Head & Medium & Tail & All \\
        \hline
        Max & 57.12 & 57.83 & 0.00 & 43.20 & 78.75 & 38.52 & 46.37 & 46.53 \\
        Mean & 48.47 & 46.91 & 0.00 & 35.57 & 78.03 & 43.75 & 58.82 & 51.98 \\
        ABMIL\cite{ilse2018attention}  & 64.40 & 60.23 & 0.00 & 46.21 & 77.78 & 40.86 & 50.54 & 48.63 \\
        G-ABMIL\cite{ilse2018attention} & 64.80 & 67.60 & 5.16 & 51.29 & 78.65 & 40.65 & 49.14 & 48.40 \\
        CLAM\cite{lu2021data} & 63.65 & 65.11 & 0.00 & 48.47 & 77.66 & 40.04 & 52.49 & 48.38 \\
        ADD-MIL\cite{NEURIPS2022_82764461} & 63.60 & 67.23 & 0.00 & 49.52 & 77.65 & 42.82 & 48.53 & 49.58 \\
        DSMIL\cite{li2021dual} & 63.22 & 72.84 & 0.00 & 52.23 & 78.09 & 41.80 & 56.45 & 50.29 \\
        DTFD\cite{zhang2022dtfd} & 63.44 & 64.89 & 0.00 & 48.30 & 76.20 & 36.38 & 49.68 & 45.24 \\
        FRMIL\cite{10640165} & 66.27 & \textbf{78.41} & 6.00 & 57.27 & 76.99 & 44.08 & 47.91 & 50.20 \\
        ILRA\cite{Xiang2023ExploringLP} & 62.91 & 77.93 & 2.58 & 55.34 & 77.62 & 44.42 & 54.74 & 51.67 \\
        TransMIL\cite{shao2021transmil} & 65.64 & 74.41 & 8.25 & 55.68 & 77.45 & 38.62 & 48.18 & 46.69 \\
        AMD-MIL\cite{ling2024agent} & 68.93 & 72.48 & 14.96 & 57.21 & \textbf{79.58} & 44.36 & 62.65 & 53.28 \\
        MDE-MIL & \textbf{70.59} & 76.47 & \textbf{28.57} & \textbf{63.02} & 79.19 & \textbf{46.23} & \textbf{63.11} & \textbf{54.54} \\
        \hline
    \end{tabular}
    \label{tab:mil_performance}
\end{table*}

\begin{table}
    \centering
    \setlength\tabcolsep{5pt} 
    \caption{Performance comparison with different components on various datasets.}
    \begin{tabular}{ccccccc}
        \hline
        \multirow{2}{*}{\textbf{Dataset}} & \multicolumn{2}{c}{\textbf{Component}} & \multicolumn{4}{c}{\textbf{Metric}} \\
        \cline{2-7}
        & Ensemble & Distillation & Head & Medium & Tail & ALL \\
        \hline
        \multirow{4}{*}{Camelyon$^{+}$-LT} 
        & none & & 68.93 & 72.48 & 14.96 & 57.21 \\
        & separate & \checkmark & 65.22 & 68.55 & 11.76& 53.52 \\
        & shared &  & 66.67 & 68.79 & 22.86 & 51.42 \\
        & shared & \checkmark & \textbf{70.59} & \textbf{76.47} & \textbf{28.57} & \textbf{63.02} \\
        \hline
        \multirow{4}{*}{PANDA-LT} 
        & none & &79.58 & 44.36 & 62.65 & 53.28 \\
        & separate & \checkmark & 78.41 & 42.64 & 59.52 & 51.42 \\
        & shared & & 79.07 & 43.07 & 63.12 & 52.41 \\
        & shared & \checkmark & \textbf{79.19} & \textbf{46.23} & \textbf{63.11} & \textbf{54.54} \\
        \hline
    \end{tabular}
    \label{tab:component_performance}
\end{table}
\begin{table}
    \centering
    \setlength\tabcolsep{3pt} 
    \caption{Performance of MDE acting as a universal adapter on various datasets.}
    \begin{tabular}{c|cccc|cccc}
        \hline
        \multirow{2}{*}{\textbf{Method}} & \multicolumn{4}{c|}{\textbf{Camelyon$^{+}$-LT (\(r\) = 35)}} & \multicolumn{4}{c}{\textbf{PANDA-LT (\(r\) = 17)}} \\
        & Head & Medium & Tail & All & Head & Medium & Tail & All \\
        \hline
        ABMIL\cite{ilse2018attention} & 64.40 & 60.23 & 0.00 & 46.21 & 77.78 & 40.86 & 50.54 & 48.63 \\
        +MDE & 65.22 & 75.93 & 17.64 & 58.68 & 79.30 & 44.75 & 59.44 & 52.96 \\
        $\Delta$ & \textcolor[rgb]{0.07,0.62,0.34}{+0.82} & \textcolor[rgb]{0.07,0.62,0.34}{+15.70} & \textcolor[rgb]{0.07,0.62,0.34}{+17.64} & \textcolor[rgb]{0.07,0.62,0.34}{+12.47} &  \textcolor[rgb]{0.07,0.62,0.34}{+1.52} & \textcolor[rgb]{0.07,0.62,0.34}{+3.89} & \textcolor[rgb]{0.07,0.62,0.34}{+8.90} & \textcolor[rgb]{0.07,0.62,0.34}{+4.33} \\
        \hline
        AMD-MIL\cite{ling2024agent}  & 68.93 & 72.48 & 14.96 & 57.21 & 79.58 & 44.36 & 62.65 & 53.28 \\
        +MDE & 70.59 & 76.47 & 28.57 & 63.02 & 79.19 & 46.23 & 63.11 & 54.54 \\
        $\Delta$ & \textcolor[rgb]{0.07,0.62,0.34}{+1.66} & \textcolor[rgb]{0.07,0.62,0.34}{+3.99} & \textcolor[rgb]{0.07,0.62,0.34}{+13.61} & \textcolor[rgb]{0.07,0.62,0.34}{+5.81} & \textcolor[rgb]{0.5,0.5,0.5}{-0.39} & \textcolor[rgb]{0.07,0.62,0.34}{+1.87} & \textcolor[rgb]{0.07,0.62,0.34}{+0.46} & \textcolor[rgb]{0.07,0.62,0.34}{+1.26} \\
        \hline
        TransMIL\cite{shao2021transmil} & 65.64 & 74.41 & 8.25 & 55.68 & 77.45 & 38.62 & 48.18 & 46.69 \\
        +MDE & 63.83 & 81.10 & 18.18 & 61.05 & 76.22 & 41.78 & 57.14 & 50.08 \\
        $\Delta$ & \textcolor[rgb]{0.5,0.5,0.5}{-1.81} & \textcolor[rgb]{0.07,0.62,0.34}{+6.69} & \textcolor[rgb]{0.07,0.62,0.34}{+9.93} & \textcolor[rgb]{0.07,0.62,0.34}{+5.37} & \textcolor[rgb]{0.5,0.5,0.5}{-1.23} & \textcolor[rgb]{0.07,0.62,0.34}{+3.16} & \textcolor[rgb]{0.07,0.62,0.34}{+8.96} & \textcolor[rgb]{0.07,0.62,0.34}{+3.39} \\
        \hline
    \end{tabular}
    \label{tab:mil_performance}
\end{table}

\subsection{Ablation Study}
\textbf{Ablation study of MDE components.}
As shown in Table \ref{tab:component_performance}, we conducted ablation studies on whether to integrate multiple branches, whether to share aggregator parameters, and whether to perform text distillation. The results demonstrate the effectiveness of the different components of MDE.

\begin{figure}[htbp]
\includegraphics[width=\textwidth]{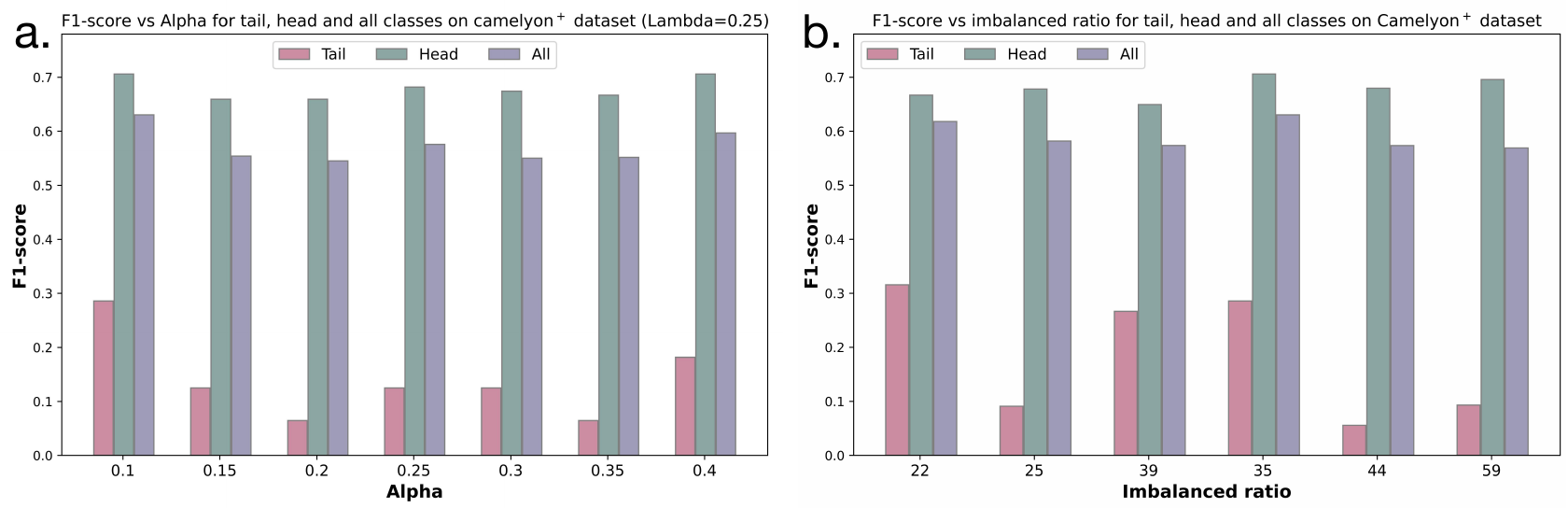}
\caption{Hyperparameter Ablation. (a) Alpha/Lambda. (b) Imbalanced ratio.
} \label{fig2}
\end{figure}

\noindent \textbf{Impact of Alpha and Lambda.}
As shown in Figure \ref{fig2}(a), we conducted experiments with different hyperparameter values. On Camelyon$^{+}$-LT, \( \alpha \) = 0.1  and  \( \lambda \) = 0.25  achieved optimal performance. On PANDA-LT, optimal performance was obtained with \( \alpha \) = 0.35  and  \( \lambda \) = 0.3.

\noindent \textbf{Effects of imbalanced ratio.}
As shown in Figure \ref{fig2}(b), experiments under different imbalance ratios reveal that increasing imbalance reduces model attention to tail classes. However, MDE mitigates this trend. 

\noindent \textbf{MDE acts as a universal adaptor.}
As shown in Table \ref{tab:mil_performance}, MDE can be adapted to various MIL frameworks to achieve better performance under long-tailed distributions. 

\section{Conclusion}
This paper proposes a multimodal distillation-driven ensemble learning method for WSI analysis using a dual-branch structure, shared-weight aggregator, and learnable prompt-based distillation mechanism. Experiments demonstrate superior performance and effective mitigation of the long-tailed distribution problem.

%
%
%
\subsubsection{\ackname}

This work was supported by the National Natural Science Foundation of China (NSFC) under Grant No. 82430062. We also gratefully acknowledge the support from the Shenzhen Engineering Research Centre (Grant No. XMHT20230115004) and the Shenzhen Science and Technology Innovation Commission (Grant No. KCXFZ20201221173207022). Additionally, we thank the Jilin FuyuanGuan Food Group Co., Ltd. for their collaboration.

\subsubsection{Disclosure of Interests.}
The authors have no competing interests to declare that they are
relevant to the content of this article.

\bibliographystyle{splncs04}
\bibliography{main}
%




\end{document}